\pdfoutput=1
\PassOptionsToPackage{dvipsnames}{xcolor}
\documentclass[11pt]{article}

\usepackage[preprint]{acl}

\usepackage{times}
\usepackage{latexsym}
\usepackage{animate}
\usepackage[T1]{fontenc}

\usepackage[utf8]{inputenc}

\usepackage{microtype}

\usepackage{inconsolata}

\usepackage{graphicx}
\usepackage{amsmath}
\usepackage{color}
\usepackage{booktabs}
\usepackage{caption}
\usepackage{subcaption}
\usepackage{enumitem}
\usepackage{multirow}
\usepackage{multicol}
\usepackage{cuted}
\usepackage{float}
\usepackage{pifont}
\usepackage{amssymb}

\usepackage{color-edits}
\addauthor{gn}{magenta}

\newcommand{\modelname}{\textsc{ViMi}}

\title{\modelname: Grounding \underline{Vi}deo Generation through \underline{M}ulti-modal \underline{I}nstruction}

\author{
Yuwei Fang\textsuperscript{1} \;\;\;\;\; Willi Menapace\textsuperscript{1} \;\;\;\;\; Aliaksandr Siarohin\textsuperscript{1} \;\;\;\;\; Tsai-Shien Chen\textsuperscript{2} \\ \bf{Kuan-Chien Wang\textsuperscript{1} \; Ivan Skorokhodov\textsuperscript{1} \; Graham Neubig\textsuperscript{3} \;  Sergey Tulyakov\textsuperscript{1}} \vspace{10pt} \\
Snap Inc.\textsuperscript{1} \quad UC Merced\textsuperscript{2} \quad  Carnegie Mellon University\textsuperscript{3} \\[1mm]
{\href{https://snap-research.github.io/VIMI/}{ \url{snap-research.github.io/VIMI}}}
}

\definecolor{darkorange}{rgb}{1.0, 0.55, 0.0}

\newif\ifdraft
\draftfalse
\ifdraft
\newcommand{\jwc}[1]{{\color{ForestGreen}[\textbf{Jackson:} #1]}}
\newcommand{\yuwei}[1]{\textcolor{darkorange}{YF: #1}}
\newcommand{\willi}[1]{\textcolor{red}{W: #1}}
\newcommand{\sergey}[1]{{\color{cyan}{S: #1}}}
\newcommand{\alex}[1]{\textcolor{blue}{A: #1}}

\else
\newcommand{\jwc}[1]{}
\newcommand{\yuwei}[1]{}
\newcommand{\willi}[1]{}
\newcommand{\sergey}[1]{}
\newcommand{\alex}[1]{}

\fi

\newcommand{\mistery}[1]{{\color{purple}{\textbf{}}}}

\definecolor{highlightcolor}{HTML}{0071BC}

\newcommand{\netf}{\mathcal{F}_{\theta}}
\newcommand{\netd}{\mathcal{D}_{\theta}}

\newcommand{\noise}{\boldsymbol{\epsilon}} 

\newcommand{\xx}{\boldsymbol{x}} 

\newcommand{\cskip}{c_\text{skip}}
\newcommand{\cin}{c_\text{in}}
\newcommand{\cout}{c_\text{out}}

\newcommand{\lossweight}{\lambda}

\newcommand{\sigmanoise}{\boldsymbol{\sigma}} 

\newcommand{\boldi}{\mathbf{I}}

\begin{document}
\maketitle
\begin{abstract}
Existing text-to-video diffusion models rely solely on text-only encoders for their pretraining. This limitation stems from the absence of large-scale multimodal prompt video datasets, resulting in a lack of visual grounding and restricting their versatility and application in multimodal integration.
To address this, we construct a large-scale multimodal prompt dataset by employing retrieval methods to pair in-context examples with the given text prompts and then utilize a two-stage training strategy to enable diverse video generation tasks within the same model. In the first stage, we propose a multimodal conditional video generation framework for pretraining on these augmented datasets, establishing a foundational model for grounded video generation. Secondly, we fine-tune the model from the first stage on three video generation tasks, incorporating multimodal instructions. This process further refines the model's ability to handle diverse inputs and tasks, ensuring seamless integration of multimodal information.
After this two-stage training process, \modelname~demonstrates multimodal understanding capabilities, producing contextually rich and personalized videos grounded in the provided inputs, as shown in Figure~\ref{fig:teaser}. 
Compared to previous visual grounded video generation methods, \modelname{} can synthesize consistent and temporally coherent videos with large motion while retaining the semantic control. Lastly, \modelname{} also achieves state-of-the-art text-to-video generation results on UCF101 benchmark.

\end{abstract}

\section{Introduction}
\begin{figure*}
\centering
\includegraphics[width=0.98\textwidth]{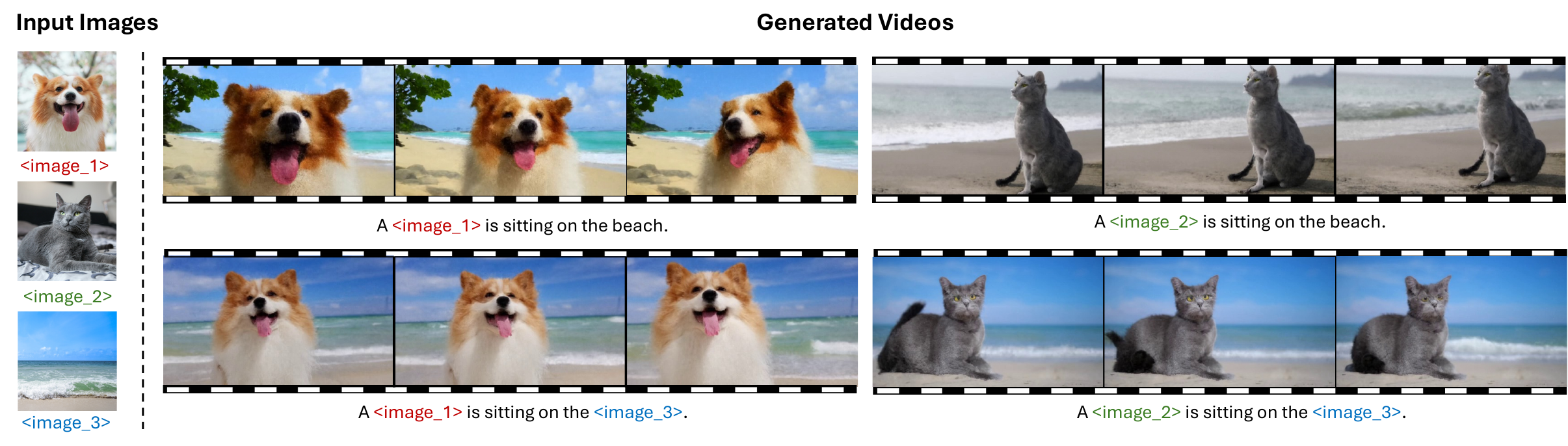}
    \caption{\textbf{Examples of ~\modelname{} for grounded video generation}. Thanks to our visual grounding during retrieval-augmented pretraining and multimodal instruction tuning, our generator can generate videos from multimodal prompts that include multiple image entities. Each multimodal prompt is displayed below the generated videos, illustrating the model's capability to integrate and interpret both textual and visual inputs effectively.} 
  \label{fig:teaser}
\end{figure*}

Recent advancements in video diffusion models have led to significant successes across various video creation tasks~\cite{singer2022makeavideo,villegas2022phenaki,zhang2023show,chai2023stablevideo,chen2023control,ceylan2023pix2video,geyer2023tokenflow}. These models have demonstrated impressive capabilities in generating high-quality videos from textual prompts~\cite{an2023latentshift,blattmann2023alignyourlatents,ge2023preserve,guo2023animatediff,he2023latent,ho2022imagenvideo,ho2022video,singer2022makeavideo,wang2023videofactory,zhou2023magicvideo,blattmann2023stable}. However, the majority of these models rely solely on text-only encoders for their diffusion-based pretraining. This limitation stems from the absence of large-scale multimodal prompt datasets, which results in a lack of visual grounding during the pretraining stage. Consequently, current models struggle to incorporate visual input effectively, restricting their versatility and application in scenarios that demand multi-modal integration.

To effectively incorporate visual input into pretrained text-to-video models, standalone image encoders are often employed to process image prompts~\cite{jiang2023videobooth,guo2023i2v,ren2024consisti2v,he2024id}. The visual embeddings generated by these encoders are then injected into the diffusion model, enabling it to handle multimodal applications. However, this approach necessitates customized model designs, leading to fragmented solutions that cannot support various tasks in a unified manner. As a result, the models lack the flexibility and generalization needed to seamlessly integrate different modalities for diverse video generation tasks.

Recently, generative pretrained multimodal language models have demonstrated robust multimodal in-context learning capabilities, showcasing their ability to process and integrate various types of input data effectively~\cite{team2023gemini,zhu2023minigpt,achiam2023gpt,liu2024visual}. Inspired by this success, we introduce a \emph{multi-modal instruction} pretraining framework \modelname~for grounded video generation. This novel framework aims to leverage the strengths of multimodal models, enhancing the ability to generate videos that are coherently grounded in both textual and visual inputs. Specifically, the training of \modelname~ consists of two stages: (1) Retrieved Augmented Pretraining; and (2) Multi-Modal Instruction Fine-Tuning. 

During the pretraining stage, we first construct a large-scale multimodal prompt dataset by employing a large-scale retrieval method to pair multimodal in-context examples with the given text prompts. The retrieved contexts from a web-scale corpus provide a rich multimodal in-context environment for model training. With these paired datasets, we can either pretrain a multimodal video generator from scratch or fine-tune an existing text-to-video generator. After this stage, the model gains the capability to understand both text-only and multimodal inputs for video generation. This establishes a foundation model for grounded video generation, capable of integrating diverse modalities into cohesive video outputs.

In the second stage, we fine-tune the model from the first stage on various video generation tasks, incorporating multimodal instructions. This fine-tuning process further refines the model's ability to handle diverse inputs and tasks, ensuring it can seamlessly integrate multimodal information. After this two-stage training process, \modelname~demonstrates enhanced multimodal understanding capabilities, producing contextually rich and personalized videos grounded in the provided inputs. This makes the model highly versatile and effective for a wide range of video generation applications.

In summary, our main contributions include:
\begin{itemize}
    \item \textbf{Novel Dataset Construction}: We are the first to use retrieval methods to build large-scale multimodal dataasets for video pretraining.
    \item \textbf{Retrieval Augmented Video Pretraining}: We propose a novel retrieval-augmented pretraining framework specifically designed for grounded video generation. 
    Our pretraining framework enables video generators to receive multi-modal prompts, instead of text-only prompts.
    \item \textbf{Instruction Tuning for Video Generation}: We introduce instruction tuning for video generation, unifying three distinct video generation tasks within a single, cohesive instruction framework. This innovative approach allows the model to flexibly handle various video generation tasks based on specific instructions.
\end{itemize}

\section{Preliminary}
\begin{figure*}
    \centering
    \begin{subfigure}{0.49\textwidth}
        \includegraphics[width=\textwidth]{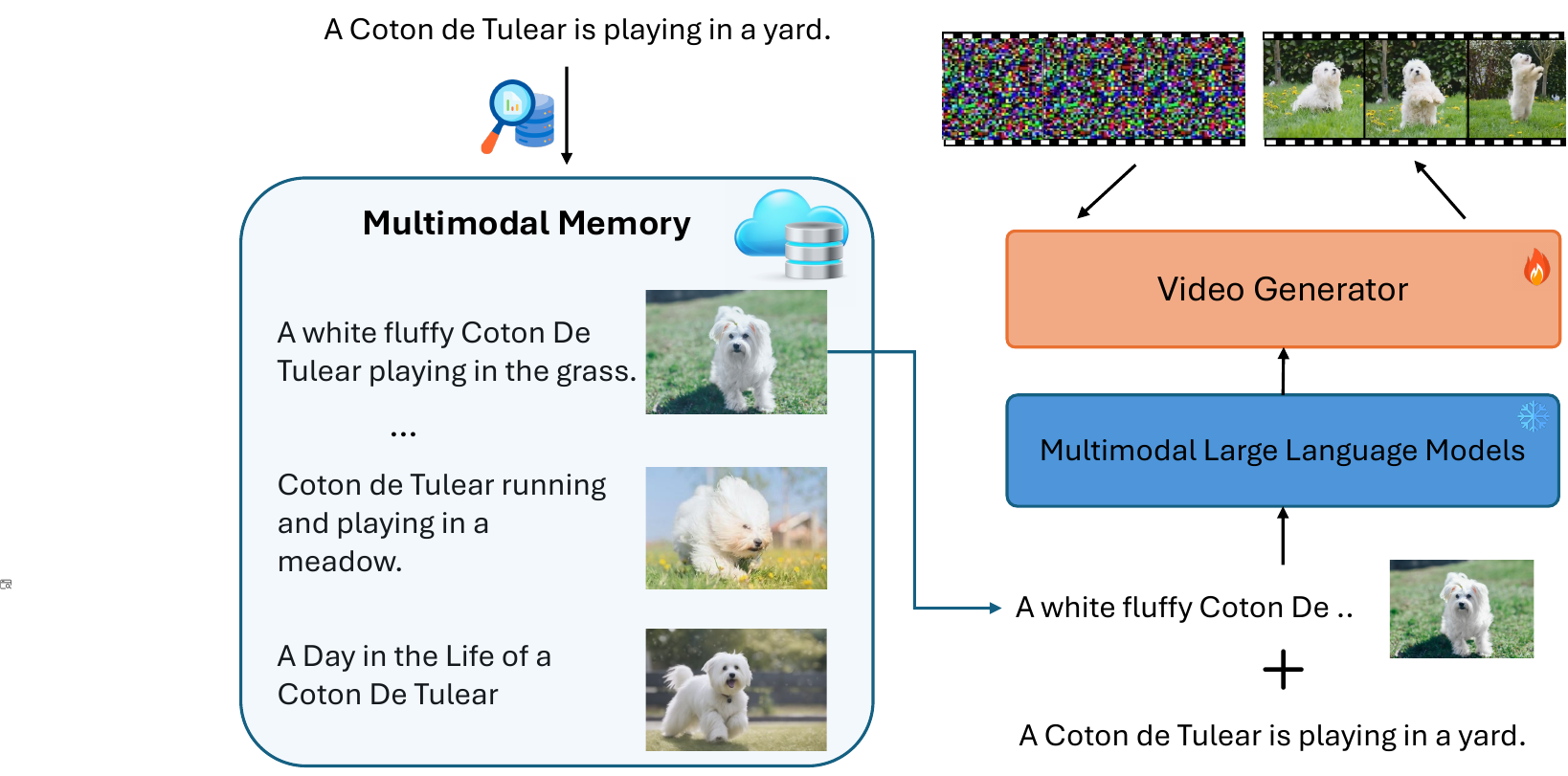}
        \caption{Retrieve-Augmented Pretraining for Videos}
        \label{fig:pretraining}
    \end{subfigure}
    \hfill
    \begin{subfigure}{0.47\textwidth}
        \includegraphics[width=\textwidth]{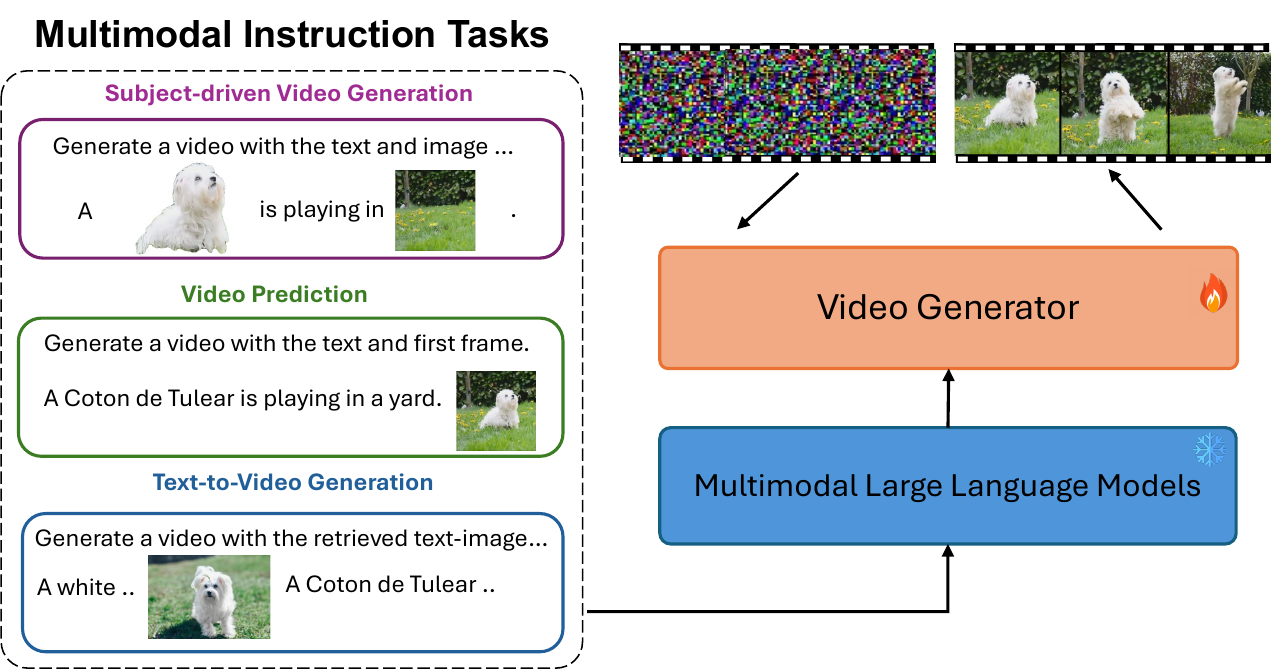}
        \caption{Multimodal Instruction Tuning for Videos}
        \label{fig:arch_instruction_tuning}
    \end{subfigure}
    \caption{\textbf{Overview of our \modelname~framework.} (a-left) We first construct a large-scale dataset by employing retrieval methods to pair multimodal in-context with given text prompts. Then we present a multimodal conditional video generation framework for pretraining on these augmented datasets. (b) We propose multimodal instruction tuning for video generation, grounding the model on customized input specified in different multimodal instructions for video generation, including subject-driven video generation, video prediction and text-to-video. By fine-tuning the model with multimodal instructions, we enable \modelname~ to generate videos that are both contextually rich and visually accurate across a wider range of tasks.} 
  \label{fig:framework}
\end{figure*}

\subsection{Text-To-Video Pretraining}
\label{sec:t2v_pretraining}
We base our work on the diffusion framework proposed by \citet{menapace2024snap}, which adapts the EDM~(\citet{karras2022elucidating}) diffusion framework to high resolution video generation.
In EDM, the forward diffusion process is characterized by a variance-exploding mechanism
$p(\xx_{\sigmanoise}|\xx) \sim \mathcal{N}(\xx, \sigmanoise^2 \boldi)$, where noise $\sigmanoise$ is gradually added to the data, causing the variance to increase over time, and $\xx_{\sigmanoise}$ represents the data at the current noise level.
The reverse process is modeled by learnable denoiser function denoted as $\netd$, which is trained using a denoising objective formulated as:
\vspace{-1mm}
\begin{equation}
  \mathcal{L}(\netd) = \mathbb{E}_{\sigmanoise, \xx, \noise} \Big[ \lossweight(\sigmanoise) ~\big\lVert \netd(\xx_{\sigmanoise}) - \xx \big\rVert^2_2 \Big]
  \label{eq:netd_loss}
  \text{,}
  \vspace{-1mm}
\end{equation}
where $\xx$ is a data sample, $\lossweight$ is a weighting function for the loss and $\noise$ is gaussian noise. Rather than learning $\netd(\xx_{\sigmanoise})$ directly, it is parametrized as:
\vspace{-1mm}
\begin{equation}
    \netd(\xx_{\sigmanoise}) = \cout(\sigmanoise) \netf\left(\cin(\sigmanoise) \xx_{\sigmanoise}\right) + \cskip(\sigmanoise)\xx_{\sigmanoise}
    \label{eq:netd}
    \text{,}
\vspace{-1mm}
\end{equation}
where $\netf$ is a neural network. By appropriately choosing scaling functions $\cout$, $\cskip$ and $\cin$ (see \citet{menapace2024snap}), the model can train optimally on high resolution videos. We employ a second order Runge-Kutta sampler to produce video samples.

\subsection{Multimodal Large Language Models}
\label{sec:mllms}

Building upon the success of Large Language Models (LLMs), Multimodal Large Language Models (MLLMs) \cite{liu2024visual,zhu2023minigpt,team2023gemini} integrate visual information from a pretrained vision encoder~\cite{radford2021learning} with an advanced LLM~\cite{touvron2023llama,jiang2023mistral}. This integration is achieved by treating visual modalities as sequences of discrete tokens. In our work, we utilize MLLMs to process and interpret multimodal in-context input data $s=(s_1, s_2, ..., s_n)$, where $s_i$ can be a signal unit, such as an image. For the image unit  $s_i$ in the prompt, a pre-trained CLIP visual encoder ViT-L/14, is used to provide the visual features $v_i = \text{Visual-Encoder}(s_i)$. The patch features $v_i$ before the last Transformer layer, combined with the text tokens, are used for MLLM encoding, formulated as:
\vspace{-1mm}
\begin{equation}
    \mathbf{C} = \text{MLLM}(\{s_1, s_2, ..., s_n\} | \mathcal{W}(v_i))
    \label{eq:mllm}
    \text{,}
\vspace{-1mm}
\end{equation}
where $\mathcal{W}$ projects $v_i$ to connect image features into the word embeddings. This approach allows the MLLM to effectively interpret a combination of textual and visual inputs, leveraging the strengths of both modalities to enhance the model's multimodal understanding and generation capabilities.

\section{Method}
\begin{figure*}
\centering
\includegraphics[width=0.95\textwidth]{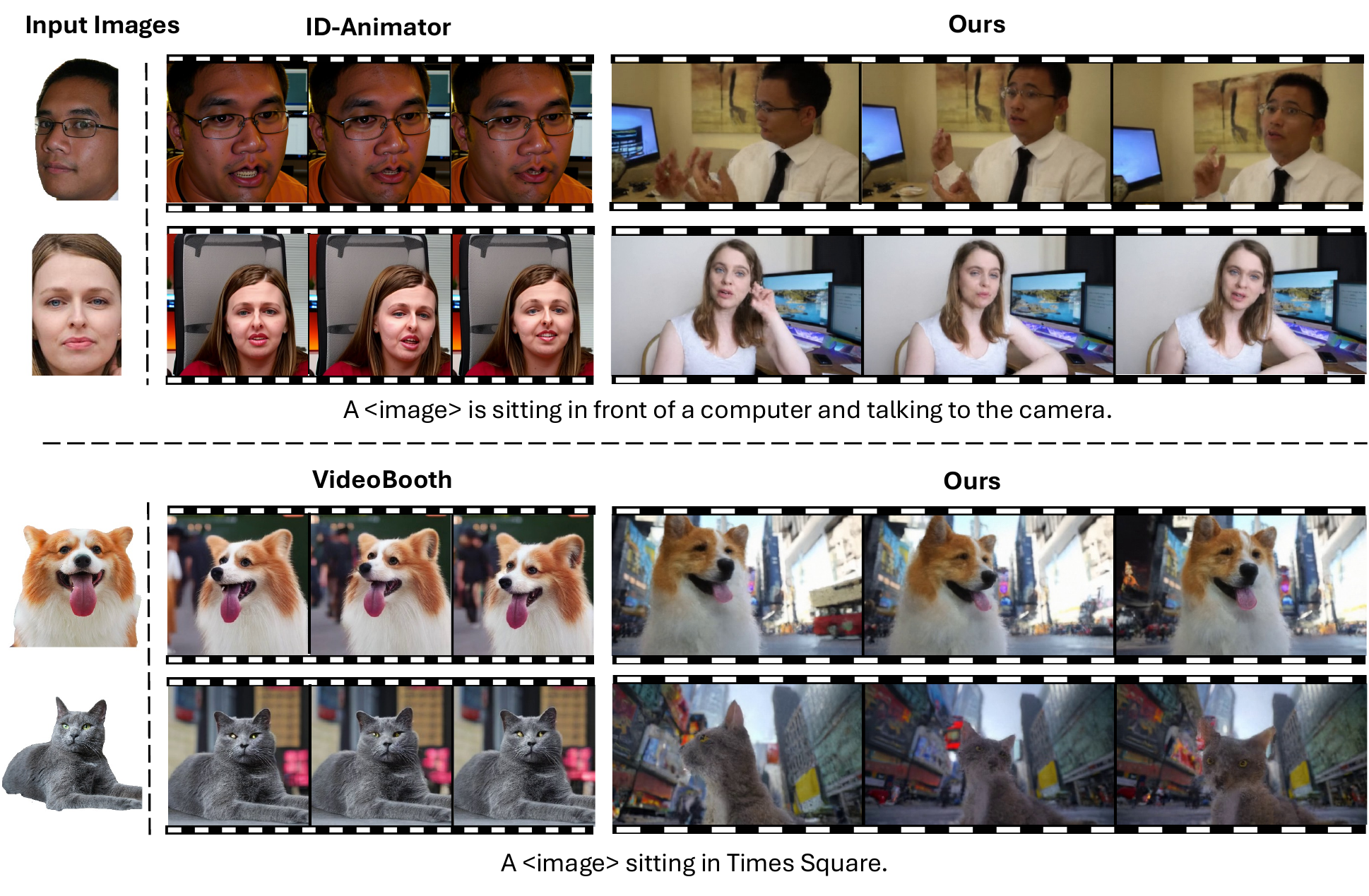}
    \caption{\textbf{Comparison of subject-driven video generation}. We compared with concurrent work ID-Animator~\cite{he2024id} for zero-shot human video generation (above) and VideoBooth~\cite{jiang2023videobooth} for general subject-driven video generation (below). Our video generator can synthesize temporally coherent videos with large motion while retaining the semantic control.}
  \label{fig:subject_driven}
\end{figure*}

We aim to generalize the video generation pretraining to the multimodal setting. Figure~\ref{fig:framework} shows the overview of our framework. Sec.~\ref{sec:rag-dataset} introduces how we construct a large-scale multimodal input-video dataset by employing retrieval methods to pair in-context examples with given text prompts. 
Sec.~\ref{sec:pretraining} presents a multimodal conditional video generation framework for pretraining on these augmented datasets, establishing a foundational model for grounded video generation. Sec.~\ref{sec:instruction-tuning} introduces the instruction finetuning stage on three video generation tasks, incorporating multimodal instructions.

\subsection{Retrieval-Augmented Multi-modal Datasets}
\label{sec:rag-dataset}

Retrieval-based methods collect relevant information to the input from an external multimodal memory $\mathcal{M}$. In our study, we use web-scale image-text pairs as our multi-modal memory for retrieval and build index into a list of key-value pairs, i.e. $\mathcal{M}=\{(k_i, v_i)\}$. Then, given the input sequence $s$, the retrieval engine $\mathcal{E}$ matches it with all keys and returns the top $K$ most similar keys to the query together with their values:
\begin{equation}
    \{(k_{i_1}, v_{i_1}), ..., (k_{i_K}, v_{i_K})\} = \mathcal{E}(s|\mathcal{M})
\end{equation}

In this work, we build the retrieval engine based on the widely used BM25 score \cite{schutze2008introduction}. We choose BM25 over dense representations due to the large scale of the retrieval datastore  and its faster speed. In our work, we construct 500M image-text pairs as our multimodal memory. Using this retrieval approach, we augment our internal text-to-video and text-to-image datasets. Specifically, we use the text caption as the query and retrieve the top-3 image-text pairs from the memory $\mathcal{M}$ for model training. These retrieved multimodal documents are then combined with the text input to form the new multimodal input, which serves as the condition for video pretraining, ensuring that the model receives contextually relevant and diverse multimodal information.

\subsection{Retrieval-Augmented Video Pretraining}
\label{sec:pretraining}
Given the retrieval-augmented multimodal input, we first concatenate the text caption $s$ with the retrieved multiodal documents to form the new multimodal input. Then, we feed this combined input into the Multimodal Large Language Models (MLLMs) to generate the multimodal conditional embedding $\mathbf{C}$:
\begin{equation}
    \mathbf{C}=\text{MLLM}(\mathcal{F}(\{(k_{i_1}, v_{i_1}), ..., (k_{i_K}, v_{i_K})\}, s)
\end{equation}
Here, $\mathcal{F}$ denotes concatenation and the embedding $\mathbf{C}$ encapsulates the rich contextual information from both the text and the retrieved multimodal data.

Following~\cite{menapace2024snap}, we use FITs~\cite{chen2023fit} as the backbone to jointly model the spatial and temporal dimensions for high-quality video generation. However, here we only use the multimodal conditioning embedding $\mathbf{C}$ to control the generation process rather than the text embeddings from T5 text encoder. We concatenate additional tokens representing the diffusion timestep, framerate and original resolution of the current input, to support variable video framerates and large differences in resolution and aspect ratios in the training data. To generate high-resolution outputs, we pretrain a cascade model consisting of a first-stage model producing $36 \times 64$px videos and a second-stage upsampling model producing $288 \times 512$px videos.

\subsection{Multimodal Instruction Tuning}
\label{sec:instruction-tuning}

After the first stage of retrieval-augmented pretraining, \modelname~can generate videos from prompts involving both text and images, leveraging the multimodal understanding capabilities of the multimodal language model. However, this initial stage primarily focuses on grounding the model in the noisy retrieved in-context input for video generation. As a result, \modelname~ may not fully utilize visual features for precise and faithful video generation. 

To address these limitations, we propose multimodal instruction tuning for video generation, grounding the model on customized input specified in different multimodal contexts for video generation. By fine-tuning the model with multimodal instructions, we enhance its ability to integrate and utilize visual features more effectively, enabling \modelname~ to generate videos that are both contextually rich and visually accurate across three video tasks, illustrated in Figure~\ref{fig:arch_instruction_tuning}.

\begin{figure}
    \includegraphics[width=0.5\textwidth]{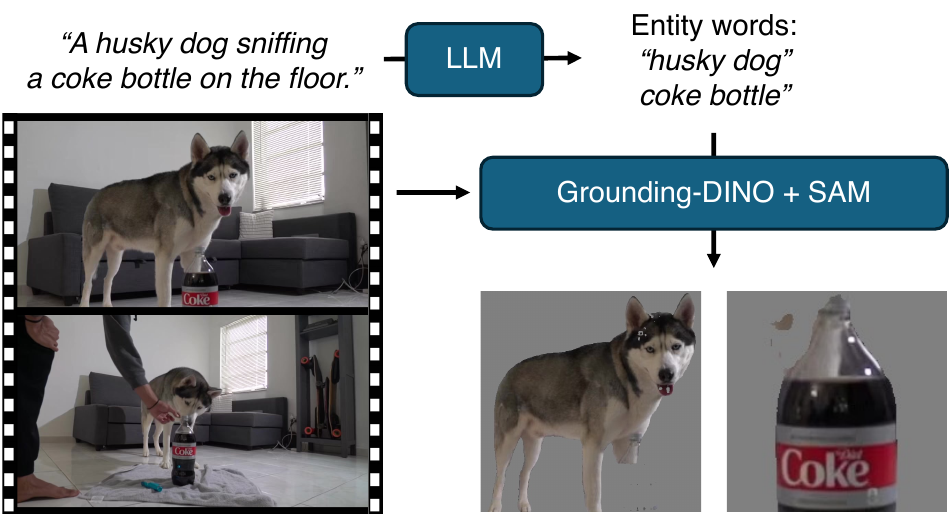}
    \caption{An overview of our data curation pipeline for subject-driven video generation.}
    \label{fig:pipeline}
\end{figure}

\vspace{-1mm}
\paragraph{Subject-Driven Video Generation}
To enhance the visual grounding capabilities for video generation, we curate a multimodal interleaved prompt composed of texts and images based on the Panda-70M dataset~\cite{chen2024panda70m}. The data curation pipeline is illustrated in Figure~\ref{fig:pipeline}.
First, we extract entity words from the text captions using Large Language Models (LLMs). For each entity, we extract the corresponding image segment using Grounding DINO~\cite{liu2023grounding} for object detection and SAM~\cite{kirillov2023segany} for image segmentation. This process ensures that each textual element has a visually grounded counterpart.
We prepend the task instruction "Generate a video with the text and image interleaved prompt." to the prompt. This curated data ensures that the model can ground specific multimodal inputs effectively and generate videos that faithfully represent the combined textual and visual information.

\paragraph{Video Prediction}
As our framework can flexibly encode multimodal prompts, we simply encode the first frame along with the text prompt with MLLMs. Following this, we generate subsequent frames based on the given multimodal prompt. To facilitate this process, we prepend the task instruction "Generate a video with the following text and first frame." to the prompt.
This approach allows the model to anchor the video generation process with a visual starting point, ensuring that the subsequent frames are coherently built upon both the initial visual and textual inputs.

\paragraph{Text-to-Video Generation}
We also use our augmented text-to-video dataset for instructed text-to-video generation. Initially, the input comprises only text. To enhance this input, we leverage retrieval methods as described in~\ref{sec:rag-dataset} to augment it with retrieved images. We prepend the task instruction "Generate a video with the retrieved text-image examples and text prompt." to the prompt, setting a clear directive for the model. This approach ensures that the model receives enriched and contextually relevant multimodal data, improving its capability to generate high-quality videos from multimodal in-context descriptions.

\begin{table}
\begin{center}
\footnotesize
\begin{tabular}{lcc}
\toprule
 & FVD $\downarrow$ & IS $\uparrow$ \\
\midrule
 CogVideo \cite{hong2022cogvideo} (Chinese) & 751.3 & 23.6 \\
 CogVideo \cite{hong2022cogvideo} (English) & 701.6 & 25.3 \\
 MagicVideo \cite{zhou2023magicvideo} & 655 & - \\
 LVDM \cite{he2023latent} & 641.8  & - \\
 Video LDM \cite{blattmann2023alignyourlatents} & 550.6 & 33.5 \\
 VideoFactory \cite{wang2023videofactory} & 410.0 & - \\
 Make-A-Video \cite{singer2022makeavideo} & 367.2 & 33.0 \\
 PYoCo \cite{ge2023preserve} & 355.2 & \textbf{47.46} \\
 VideoPoet \cite{kondratyuk2023videopoet} & 355 & 38.4 \\
 W.A.L.T \cite{gupta2023photorealistic} & 258.1 & 35.1 \\
 Lumiere \cite{bar2024lumiere} & 332.5 &  37.5 \\
 Snap Video ($288\times288$ px) &  260.1 & 38.89 \\
 Snap Video ($512\times288$ px) &  200.2 & 38.89 \\
\midrule
\modelname{} ($288\times288$ px) &   262.5 & 35.6 \\
 \modelname{} ($512\times288$ px) &   \textbf{193.7} & 35.6 \\
\bottomrule
\end{tabular}
\end{center}
\caption{Zero-shot evaluation results on UCF101 \cite{soomro2012ucf}.}
\label{table:evaluation_ucf101}
\end{table}
\begin{figure*}
    \centering
    \includegraphics[width=0.95\textwidth]{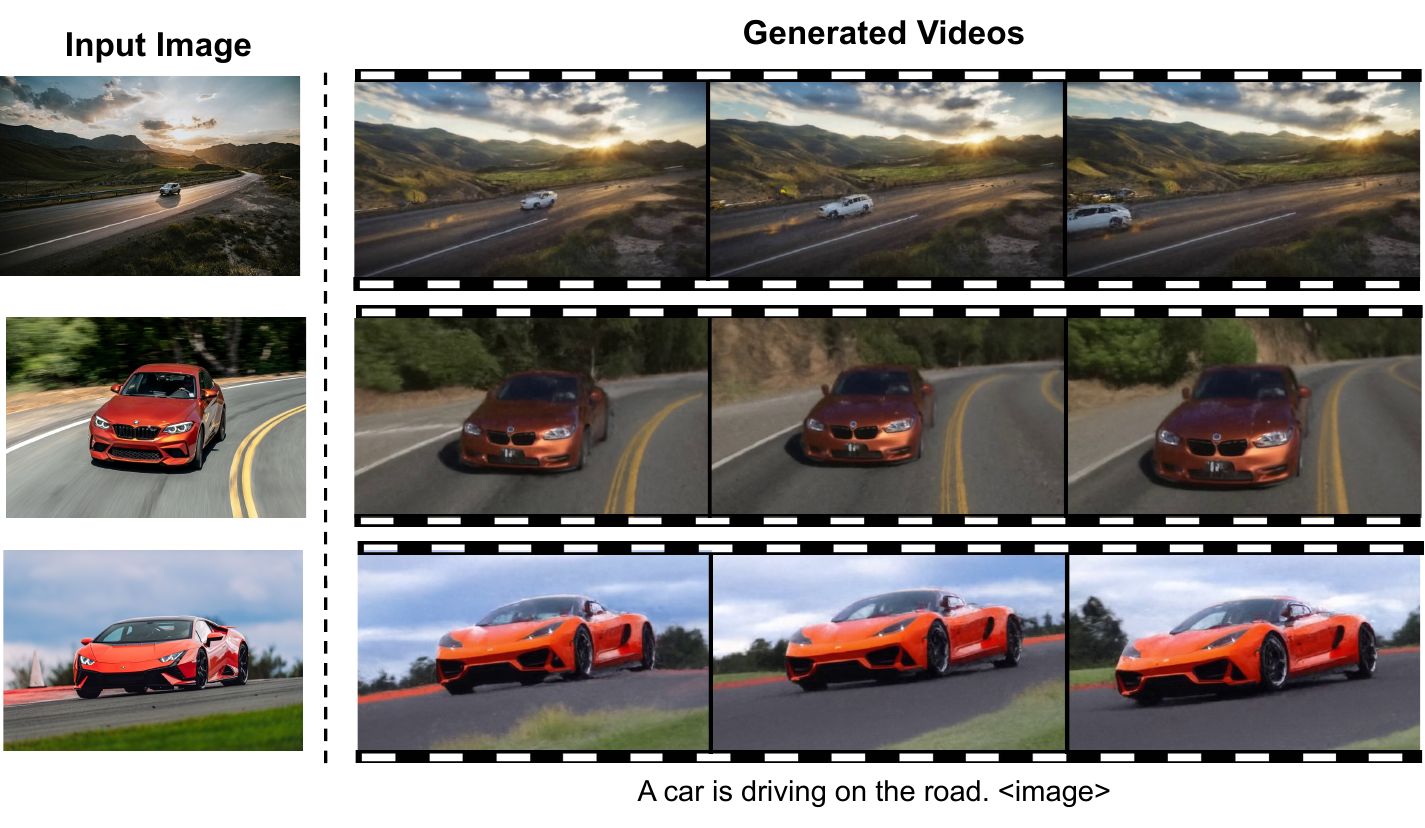}
    \caption{\textbf{Examples of Video Prediction results. \yuwei{TODO: better examples}}}
  \label{fig:prediction}
\end{figure*}

\section{Experiments}
In this section, we evaluate \modelname{} against baselines and ablate the model design components. Sec.~\ref{sec:details} introduces our implementation details. Sec.~\ref{sec:quantitative} shows our  results in three different evaluation settings: (1) general text-to-video generation; (2) subject-driven video generation; and (3) video prediction. Sec.~\ref{sec:ablation} shows ablations of our framework.

\subsection{Implementation Details}
\label{sec:details}

We use an internal licensed dataset of images and videos, each paired with a corresponding text caption.
We use the retrieval methods of section~\ref{sec:rag-dataset} to augment multimodal in-context examples for pretraining and instruction tuning. We use UCF-101~\cite{soomro2012ucf}, a video dataset from 101 action categories, for general text-to-video evaluation. We use human subjects from the CelebA~\cite{liu2015deep} and general subject from the Dreambooth~\cite{ruiz2023dreambooth} for qualitative comparison. Training and evaluation details are in Appendix~\ref{app:training} and ~\ref{app:evaluation}.

\subsection{Results}
\label{sec:quantitative}

\paragraph{Zero-shot Text-to-Video Evaluation}
We generate 10,000 videos \cite{wang2023videofactory,blattmann2023alignyourlatents} sampling classes with the same distribution as the original UCF-101 dataset. We produce a text prompt for each class label \cite{ge2023preserve} and compute FVD \cite{unterthiner2018towards} and Inception Score \cite{salimans2016improved}. Table~\ref{table:evaluation_ucf101} shows our competitive performance to previous state-of-the-art text-to-video generators in both FVD and IS metrics. We achieve the best FVD score of 193.7 which we attribute to our visual grounding during pretraining. 

\paragraph{Zero-shot Subject-driven Video generation}

Figures~\ref{fig:teaser} and~\ref{fig:subject_driven} show our results for subject-driven video generation. Compared to VideoBooth~\cite{jiang2023videobooth}, our generator can handle multimodal prompts that include multiple image entities, as illustrated in Figure~\ref{fig:teaser}. We also compared our model with the concurrent work ID-Animator~\cite{he2024id} for zero-shot human identity preservation generation in Figure~\ref{fig:subject_driven}. Overall, our video generator can not only ground on the visual input but also synthesize temporally coherent videos with large motion while retaining semantic control.

\paragraph{Video Prediction} As shown in Figure~\ref{fig:prediction}, \modelname{} can also generate videos conditioned on a single image, thanks to our unified multimodal instruction tuning stage. We first append the `<image>’ token after the text prompt and use MLLMs to encode this multimodal prompt for video prediction.

\subsection{Ablation Study}
\label{sec:ablation}

\paragraph{Effectiveness of retrieval-augmented pretraining}

\begin{figure}
    \centering
    \begin{subfigure}{0.5\textwidth}
        \includegraphics[width=\textwidth]{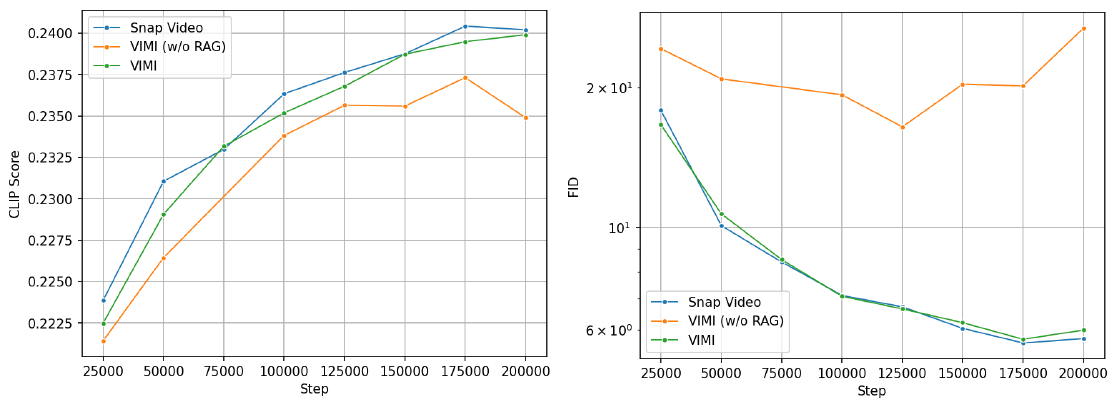}
        \caption{Effectiveness of retrieval-augmented pretraining.}
        \label{fig:ablate_pretraining}
    \end{subfigure}
    \begin{subfigure}{0.5\textwidth}
        \includegraphics[width=\textwidth]{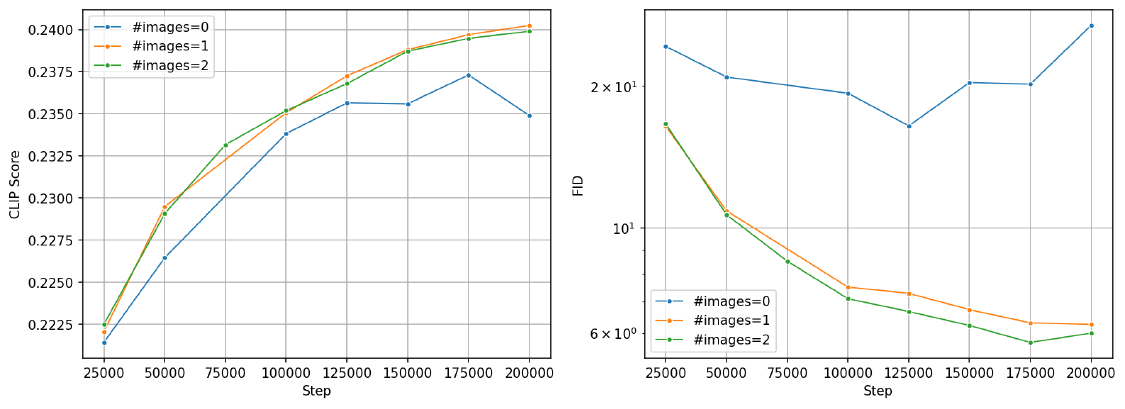}
        \caption{Effectiveness of the number of retrieved images.}
        \label{fig:ablate_retrieval}
    \end{subfigure}

    \caption{\textbf{Ablation of retrieval-augmented pretraining}. (a) shows the evaluations of retrieval-augmented pretraining on our validation set for CLIP similarity and FID metrics. We denote \modelname{} without retrieval augmented pretraining as \modelname{} (w/o RAG). (b) shows the results of pretraining with different numbers of retrieved images.}
  \label{fig:ablation}
\end{figure}

Figure~\ref{fig:ablate_pretraining} shows the evaluations of retrieval-augmented pretraining on our validation set for CLIP similarity and FID metrics. We denote \modelname{} without retrieval augmented pretraining as \modelname{} (w/o RAG).
We use Snap Video~\cite{menapace2024snap} with text encoders T5-11B as another baseline.
The results indicate that using multimodal large language models as the encoding leads to unstable model training. Specifically, the FID results converge slowly and do not decrease after 125K pretraining steps. In contrast, with retrieval augmented pretraining, \modelname{} shows faster convergence and more stable training.
After 200K pretraining steps, using a multimodal large language model as the encoder demonstrates performance comparable to Snap Video~\cite{menapace2024snap}. This highlights the effectiveness of our retrieval augmented pretraining approach in stabilizing training and improving the overall performance of the video generation model.

\paragraph{Effectiveness of the number of retrieved Images}
Figure~\ref{fig:ablate_retrieval} shows the results of pretraining with different numbers of retrieved images. We set K to be up to 2 for ablation studies, primarily considering the multimodal sequence length. We observe that using only one retrieved image stabilizes the model training. Increasing K to 2 provides further stable improvements in the early pretraining stage. After 200K pretraining steps, the model converges to comparable evaluation results for both settings. Given our aim to support multi-subject generation, we use K=2 for the pretraining. This choice balances the need for rich contextual information with the practical constraints of sequence length, ensuring stable and effective training.

\vspace{-2mm}
\paragraph{Effectiveness of multimodal instruction tuning}

\begin{figure*}
\centering
\includegraphics[width=0.98\textwidth]{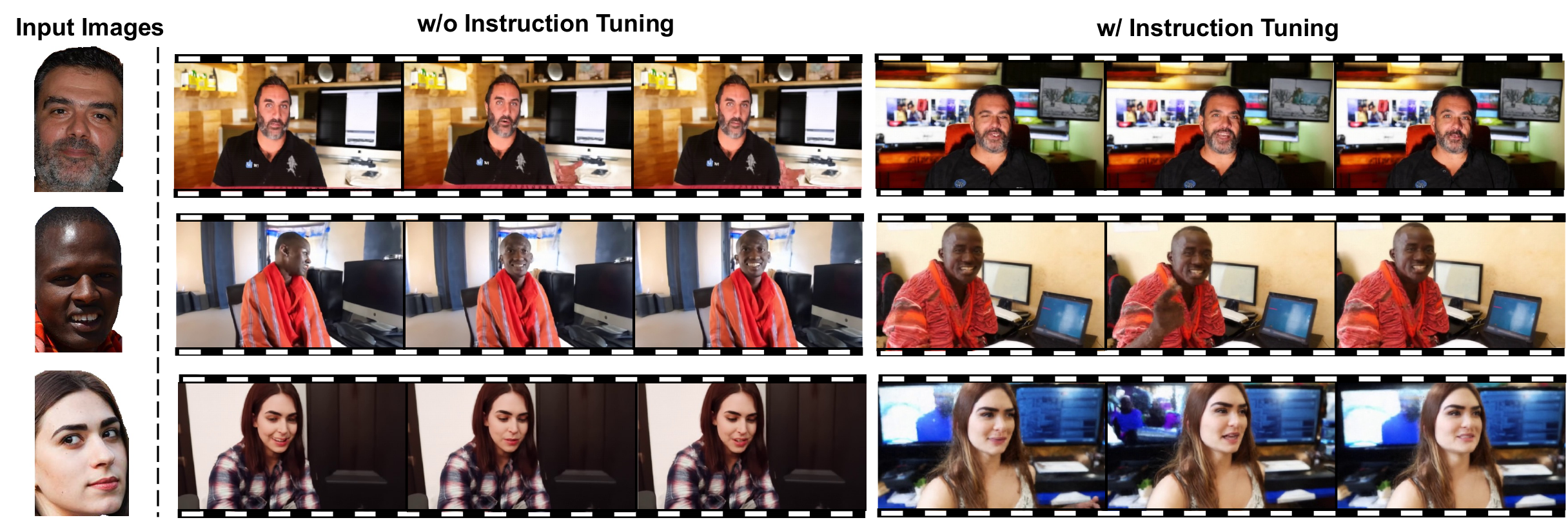}
    \caption{\textbf{Ablation of multimodal instruction tuning}.
    We compare \modelname{} with a variant finetuned only on subject-driven data during the second stage (``w/o Instruction Tuning'').
    We use the prompt ``A <image> is sitting in front of a computer and talking to the camera.''. \modelname{} achieves better semantic alignment and identity preservation.}
  \label{fig:ablation_instruction}
\end{figure*}

For the second stage, if we fine-tune only on subject-driven data (denoted as ``w/o Instruction Tuning''), \modelname{} can also generate videos from multimodal interleaved prompts. To evaluate the effectiveness of unified multimodal instruction tuning, we compare this variant of \modelname{} in subject-driven generation tasks after fine-tuning for the same 100K steps. Figure~\ref{fig:ablation_instruction} shows that multimodal instruction tuning preserves identity better and follows instructions more accurately. We attribute this improvement to the more diverse fine-tuning tasks provided by multimodal instruction tuning.

\section{Related Work}


\paragraph{Video Generation}
Diffusion models are now the standard methodology for both image~\cite{ho2020denoising,nichol2021improved,rombach2022high,song2020denoising} and video generation~\cite{an2023latentshift,blattmann2023alignyourlatents,ge2023preserve,guo2023animatediff,he2023latent,ho2022imagenvideo,ho2022video,singer2022makeavideo,wang2023videofactory,zhou2023magicvideo,blattmann2023stable}.
Early video diffusion models use the U-Net~\cite{ronneberger2015unet} for the video generation task. \citet{ho2022video} showed that jointly training on image and video data can improve text conditioned video generation greatly. 
Make-A-Video \cite{singer2022makeavideo} proposed to build on text-to-image models with novel and effective spatial-temporal modules. Video LDM \cite{blattmann2023alignyourlatents} adopts a latent diffusion paradigm where a pre-trained latent image generator and latent decoder are finetuned to generate temporally coherent videos.
Most recently, diffusion transformer~\cite{Peebles2022DiT} has been widely adopted for video generation. Latte~\cite{ma2024latte} proposes a latent diffusion transformer, which adopts a video Transformer as the backbone. 
W.A.L.T~\cite{gupta2023photorealistic} uses a transformer-based method for latent video diffusion models and a window attention architecture tailored for joint spatial and spatiotemporal generative modeling. Snap Video~\cite{menapace2024snap} replaced U-Nets with efficient
transformer-based FITs~\cite{chen2023fit} and scaled to billions of
parameters. However, these existing works are still limited by the use of text encoders like T5 or the CLIP Text encoder, which lack visual grounding in the pretraining phase. In our work, we propose to utilize multimodal large language models to encode multimodal inputs for video generation, addressing the limitations by integrating visual grounding into the pretraining process. 

\paragraph{Retrieval Augmented Multimodal Pretraining}
Retrieval augmentation has shown significant promise, particularly in language models. Initial work ~\cite{lewis2020retrieval,guu2020retrieval} demonstrated how incorporating external knowledge into a language model can enhance its performance.
This is achieved by first retrieving documents relevant to the input text from an external memory, and then integrating these retrieved documents with the input for improved modeling~\cite{hashimoto2018retrieve,karpukhin2020dense,borgeaud2022improving}. Beyond language models, recent studies have explored retrieval techniques for image generation~\cite{blattmann2022retrieval,sheynin2022knn,sarto2022retrieval,ramos2023smallcap,chen2022re}.
For instance, KNN-Diffusion~\cite{sheynin2022knn} used retrieval methods to search for k-Nearest-Neighbors images, facilitating the training of a small and efficient text-to-image diffusion model. RA-CM3~\cite{yasunaga2022retrieval} was the first multimodal model capable of retrieving and generating both text and images using autoregressive models. 
Additionally, Re-Imagen~\cite{chen2022re} employed an external multimodal knowledge base to retrieve relevant image-text pairs, using them as references for a diffusion model to generate images.
In contrast to these works, our approach is the first to uses retrieval methods to augment text-video datasets, formalizing multimodal input-video pairs for video pretraining.

\vspace{-1mm}
\paragraph{Multimodal Instruction Tuning}
Instruction tuning was first proposed to finetune a large language model with instructions to improve its zero-shot learning performance on unseen tasks~\cite{wei2021finetuned,chung2024scaling}. Inspired by its success in language domain, instruction tuning was also introduced in the vision generation domain~\cite{yu2023scaling,sun2023generative,liu2024visual,hu2024instruct}.
For instance, CM3Leon~\cite{yu2023scaling} utilized the CM3 multimodal architecture~\cite{aghajanyan2022cm3}, demonstrating the substantial benefits of scaling up and tuning on more diverse instruction-style data. Emu2~\cite{sun2023generative} demonstrated the in-context learning capabilities of large multimodal models with a unified autoregressive objective.
More recently, Instruct-Imagen \cite{hu2024instruct} introduced multi-modal instruction for image generation by fine-tuning a pre-trained text-to-image diffusion model with a two-stage framework.
In our work, we are the first to propose instruction tuning for video generation, by unifying three distinct video generation tasks within a single, cohesive instruction framework. By leveraging instruction tuning, we aim to enhance the model's ability to interpret and execute a wide range of video generation instructions, thereby improving its performance and applicability in diverse contexts.

\section{Conclusion}

In this work, we first construct a multimodal prompt dataset for video pretraining using retrieval methods. We then propose a two-stage training strategy to enable diverse video tasks within the same model. For the first stage, we introduce a multimodal conditional video generation framework for pretraining on these augmented datasets, establishing a foundational model for grounded video generation. In the second stage, we fine-tune the model from the first stage on three video generation tasks, incorporating multimodal instructions.
Our experiments demonstrate the effectiveness of retrieval-augmented pretraining and the use of multimodal instruction tuning. We hope this approach opens up new opportunities for video pretraining, such as building large-scale multimodal datasets for pretraining, utilizing stronger multimodal large language models for encoding, and employing instruction tuning for diverse video tasks.
\section{Limitations}
Firstly, similar to subject-driven image generation models, our video generator sometimes struggles to produce accurate and faithful videos. To improve visual quality, future work will focus on utilizing stronger multimodal large language models, diffusion tranformers and jointly fine-tuning these models.
Secondly, due to memory and training constraints, we only experimented with two context examples and displayed at most two image entities for multi-subject-driven generation. Extending this work to support any-subject video generation will be a goal for future research.
Thirdly, our current results are based on qualitative evaluation. Developing comprehensive evaluation methods for grounded video generation, such as any-subject-driven video generation, will be crucial for building a visually grounded video generator.

\section{Ethical Considerations}
Like all generative AI advancements, visually grounded video generation models raise important ethical considerations, such as the creation of misleading or false information and bias. Developers and researchers should consider safeguards to address these issues such as evaluating datasets, and adding watermarks or other identification mechanisms. It is important to consider the societal impacts and work towards solutions that balance innovation with social responsibility. 

\bibliography{anthology,main}
\bibliographystyle{acl_natbib}

\clearpage
\appendix

\section{Training details}
\label{app:training}
For pretraining, we can either start from scratch or initialize the weights of the model from existing text-to-video generators. In our work, we initialize the FIT weights from ~\cite{menapace2024snap}. We keep its parameters frozen for 30,000 steps to stabilize the initial training phase, and then fine-tune the entire model for an additional 100,000 steps. In the second stage, we fine-tune the model starting from the weights obtained in the first stage for 30,000 steps. We use a learning rate of $5e^{-3}$, a cosine learning schedule, and a total batch size of 256 videos and 256 images.

\section{Evaluation Protocol}
\label{app:evaluation}
We evaluate our method against baselines by following the protocols in \cite{singer2022makeavideo,ge2023preserve,wang2023videofactory,blattmann2023alignyourlatents,zhou2023magicvideo,luo2023videofusion} for zero-shot evaluation on the UCF-101 \cite{soomro2012ucf}. We generate 16 frames videos in $512 \times 288$px resolution at 24fps. To validate the effectiveness of pretraining, ablations are performed in $64 \times 36$px resolution using the first-stage model only, and compute FID \cite{heusel2017advances}, FVD \cite{unterthiner2018towards} and CLIPSIM \cite{wu2021godiva} metrics against the test set of our internal dataset on 50k generated videos.

\section{Inference}
\label{sec:inference}

We produce video samples from gaussian noise and user-provided conditioning information using the deterministic sampler of \cite{karras2022edm} and two-stage cascade. We use 256 sampling steps for the first-stage and 40 for the second-stage model, and employ classifier free guidance \cite{ho2022classifierfree} to improve text-video alignment.

\end{document}